\documentclass{article}

\usepackage[preprint,nonatbib]{nips_2018}
\usepackage[shortlabels]{enumitem}

\usepackage[utf8]{inputenc} 
\usepackage[T1]{fontenc}    
\usepackage{hyperref}       
\usepackage{url}            
\usepackage{booktabs}       
\usepackage{amsfonts}       
\usepackage{nicefrac}       
\usepackage{microtype}      
\usepackage{enumitem}
\usepackage{graphicx}
\usepackage{color}
\usepackage{subcaption}

\title{Predicting Poverty Level from Satellite Imagery using Deep Neural Networks
}

\author{
  Varun Chitturi \\
  Mission San Jose High School\\
  \texttt{varunchitturi@icloud.com} \\
  \AND
  Zaid Nabulsi \\
  Stanford Univeristy\\
  \texttt{znabulsi@cs.stanford.edu} \\
}

\begin{document}

\maketitle

\begin{abstract}
Determining the poverty levels of various regions throughout the world is crucial in identifying interventions for poverty reduction initiatives and directing resources fairly. However, reliable data on global economic livelihoods is hard to come by, especially for areas in the developing world, hampering efforts to both deploy services and monitor/evaluate progress. This is largely due to the fact that this data is obtained from traditional door-to-door surveys, which are time consuming and expensive. Overhead satellite imagery contain characteristics that make it possible to estimate the region's poverty level. In this work, I develop deep learning computer vision methods that can predict a region's poverty level from an overhead satellite image. I experiment with both daytime and nighttime imagery. Furthermore, because data limitations are often the barrier to entry in poverty prediction from satellite imagery, I explore the impact that data quantity and data augmentation have on the representational power and overall accuracy of the networks. Lastly, to evaluate the robustness of the networks, I evaluate them on data from continents that were absent in the development set. 

\end{abstract}

\section{Introduction}
The Sustainable Development Goals lists its first of 17 goals as ending all forms of poverty globally \cite{one}. However, despite this being a foremost goal of the United Nations, there are still extensive regions across the world where extreme levels of poverty are endemic. Additionally, advancement towards that goal is impeded by a consistent lack of data regarding key economic indicators, especially in the developing world. Having accurate measures of poverty helps for two main reasons:

\begin{enumerate}
    \item Accurate measurements of poverty level allows philanthropic agencies and governments to identify where resources and interventions are needed, and helps guide the direction of financial aid.
    \item Frequent and reliable data on poverty levels and distribution allows agencies to better track progress on the Sustainable Development Goals.
\end{enumerate}

Currently, poverty is formally calculated by numerous philanthropic agencies including the World Bank. One of the reasons why data on poverty is sparse in the developing world is because it is infrequently collected due to the high cost associated with on-the-ground surveys. For instance, there is an average of more than four years between national wealth surveys in a majority of African countries \cite{two}. Furthermore, at current rates of these surveys, a household in the African continent will appear less than once every 1000 years on average, which is roughly 100 times scarcer than a household in the United States \cite{three}. In order to increase the reliability of wealth data in regions like countries in Africa through these surveys, an infeasible amount of capital will need to be spent. Thus, accurately and efficiently evaluating the level of need of the developing regions of the word and in turn directing resources accordingly necessitates a method that can assess critical lifestyle metrics without these costly door-to-door surveys.

Recent advancements in deep learning present an exciting opportunity for application to poverty prediction. More specifically, both daytime and nighttime satellite imagery of regions can be used to estimate poverty in certain regions. \cite{three} Deep learning has been a main factor behind recent breakthroughs in numerous computer vision tasks such as image classification, segmentation, and object detection \cite{four, five, six, seven}. However, while the emergence of deep learning has been a boon for countless areas, training an accurate neural network requires vast amounts of labeled data \cite{eight, nine}, a luxury that the intersection between satellite imagery and poverty data does not have. While there are no comprehensive datasets with labels that currently exist for predicting poverty from satellite imagery, there does exist a plethora of satellite imagery spanning the entire world, which can be paired with the reliable poverty data that does exist.

In this paper, I test the hypothesis that deep learning can leverage satellite imagery to reliably predict the poverty level of a region. I assemble a dataset of 88,386 images from 44,193 cities spanning Africa, South America, Asia, Europe, and the Caribbean. For each city, I obtain a daytime satellite image, a nighttime satellite image, and the city’s wealth index. I then train deep neural networks (DNNs) to predict a city’s wealth index, given a satellite image. I leverage techniques such as data augmentation, pretraining and transfer learning, regularization, and cross validation in developing and evaluating the networks. I evaluate the networks using cross-country and within-country wealth data, and also explore the impact of data quantity on the representational power of the networks.

\section{Related Work}

Deep learning has been applied to satellite imagery in the context of structure classification \cite{xia2018dota}, map segmentation \cite{wang2015deep} \cite{audebert2017joint}, and pattern identification \cite{albert2017using}. Using satellite imagery to predict poverty levels of cities is not a new task. Previous work has explored predicting humanitarian indicators using satellite imagery. \cite{kim2006satellite} uses satellite imagery to predict $NO_x$ emissions, and \cite{wang2012poverty} \cite{Jean790} \cite{nips_2017_workshop_stefano} utilize Nightlight and RGB satellite imagery to predict poverty outcomes with promising results.

Furthermore, deep learning has been used to infer both spatial and temporal differences in local-level economic well-being using multiple sources of satellite imagery \cite{three}. Jean etl. Al. use CNNs to predict poverty from high resolution satellite imagery of terrain10 and Perez et. al. and Xie et. al. utilize transfer learning and multi-modal networks to predict poverty from low resolution satellite imagery \cite{eleven, twelve}. More recently, Yeh et. al. presents the utility of satellite imagery for research and policy, demonstrating their potential to create a wealth map for the entire world \cite{three}. Interestingly, Uzkent et. al. presented an approach pairing satellite imagery with geolocated Wikipedia articles to accurately predict the wealth index of a region \cite{thirteen}. Lastly, generative models have been used to discern the distribution of satellite images and identify forgery \cite{fourteen}.

To the best of my knowledge, there have been no prior work to this date that explores the impact of data quantity on deep neural networks, or evaluates such networks on data from different continents than they were trained on. Furthermore, to the best of my knowledge, there have been no prior works that explore both nighttime and daytime imagery.

\section{Methods}

I developed DNN to predict the wealth index of a region given an overhead satellite imagery. I explore DNNs that take in as input both nighttime images or daytime images. Formally, the task of poverty prediction can be defined as predicting $Y$, which is the wealth index, given a daytime satellite image, $X_{Day\_Image}$, or nighttime satellite image, $X_{Nigttime\_Image}$. The DNN will output $\hat{Y}$, the predicted wealth index of the region. Because the wealth index, $Y$, is a continuous number in contrast to a categorized , this task is framed as a regression. All in all, the goal is to predict some $Y$ by training a DNN to predict  $Y | X_{Nigttime\_Image}$ or  $Y | X_{Day\_Image}$.

\subsection{Dataset}
\begin{figure}[t]
    \centering
    \includegraphics[width=5in]{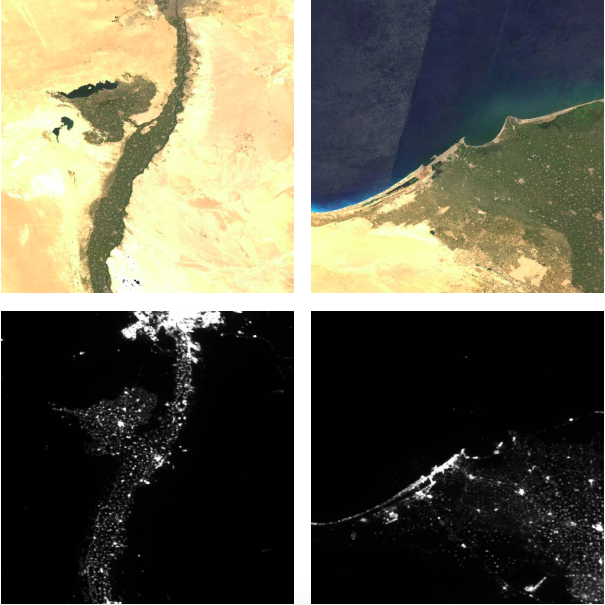}
    \caption{Sample satellite images, obtained randomly from the development dataset. On top, we see the daytime satellite images, with the corresponding nighttime images of the same regions shown directly below.}
    \label{fig:night_day}
\end{figure}

I obtain my development and evaluation satellite images from the Google Earth Engine API \cite{fifteen}. Daytime satellite imagery is obtained from the LANDSAT satellite \cite{sixteen} and nighttime satellite imagery is obtained from the VIIRS satellite \cite{seventeen}. To ingest the images, I use the Google Earth Engine API to obtain a daytime and a nighttime satellite image for 44,193 points that I had poverty levels for (see Ground Truth). For each of these points, I obtain the closest satellite image to the geographic coordinate of the wealth index. All in all, I collected 88,386 satellite images. Examples of both daytime and nighttime images are presented in Figure \ref{fig:night_day}.

All satellite images had initial spatial dimensions of 890x890 pixels but were downsampled to 256x256 pixel images. The nighttime images were single-channel (intensity) grayscale images, while the daytime images were 3-channel RGB images.

I divided the initial images into training, tuning, and testing sets with the 80:10:10 split. The splits were executed on the country level to ensure that data from the same country is not in more than one set. Furthermore, the splits were consistent across daytime and nighttime images meaning that daytime and nighttime images of the same region were in the same split, ensuring consistency in reporting metrics across daytime and nighttime experiments.

All evaluation is done on held-out test regions that were not used in model development. This mimics the real-world usage of such a network, where the network is expected to make predictions on regions unseen during training.

\subsection{Poverty Ground Truth Labels}

\begin{figure}[t]
    \centering
    \includegraphics[width=6in]{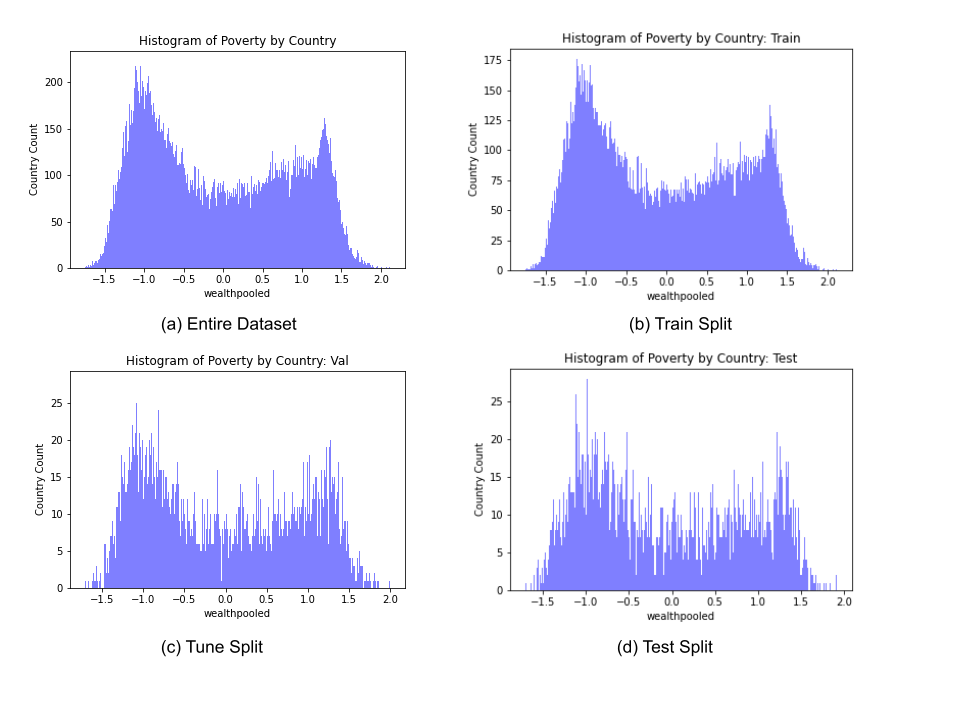}
    \caption{Distribution of wealth measurements in the entire dataset (across all splits), and for  each split. The values are normalized and range from -2 to 2. Note that the data is bi-modal, with a median close to 0. Furthermore, note that the distributions are consistent across data splits. }
    \label{Poverty_dataset}
\end{figure}

The ground truth data is the poverty level of a region, given as a wealth index. I obtain this wealth index data from the UN World Bank datasets \cite{eighteen} for a total of 44,193 coordinates. I then normalize the wealth data, ending up with numbers in the interval [-2, 2] ,where higher (less negative or more positive) numbers indicate greater wealth. Zero indicates the median wealth index in the world, negative numbers indicate below average wealth, and positive numbers indicate above average wealth. Because the wealth index is a continuous number, this formulation is framed as a regression task. The distribution of the wealth indices across the splits of my dataset is shown in Figure  \ref{Poverty_dataset}. Table \ref{tab:countries} shows some key statistics of the wealth indices on the country level.

Note that in obtaining the wealth indices' ground truth data, in order to protect the anonymity of each interview subject, a 5km jitter (error/uncertainty) was added to every coordinate in a rural area, and 2km on coordinates in urban areas.

\begin{table}[]
\begin{tabular}{lrrr}
\toprule
 country                          &     average &     median &   variance \\
\midrule
 Albania                          &  1.23173    &  1.26679   &  0.0524784 \\
 Armenia                          &  1.22638    &  1.27885   &  0.0379816 \\
 Bangladesh                       & -0.310995   & -0.409245  &  0.395392  \\
 Bolivia                          &  0.509215   &  0.763213  &  0.653936  \\
 Burkina Faso                     & -0.480204   & -0.666471  &  0.345656  \\
 Burundi                          & -0.87925    & -1.08369   &  0.320024  \\
 Cambodia                         & -0.162819   & -0.399031  &  0.48493   \\
 Cameroon                         & -0.415385   & -0.486957  &  0.493745  \\
 Central African Republic         & -1.17838    & -1.16831   &  0.0453396 \\
 Comoros                          &  0.308073   &  0.350365  &  0.249185  \\
 Cote d'Ivoire                    &  0.00953811 &  0.0686464 &  0.393958  \\
 Democratic Republic of the Congo & -0.918128   & -1.20888   &  0.425719  \\
 Dominican Republic               &  0.773491   &  0.878884  &  0.299624  \\
 Egypt                            &  1.10081    &  1.20375   &  0.130868  \\
 Ethiopia                         & -0.925519   & -1.23838   &  0.427344  \\
 Gabon                            &  0.289272   &  0.505296  &  0.509286  \\
 Ghana                            & -0.00517709 &  0.0100654 &  0.523037  \\
 Guinea                           & -0.458707   & -0.706828  &  0.525901  \\
 Guyana                           &  0.417429   &  0.650645  &  0.513695  \\
 Haiti                            & -0.277625   & -0.428153  &  0.483068  \\
 Kenya                            & -0.426766   & -0.577282  &  0.471848  \\
 Kyrgyz Republic                  &  1.21326    &  1.25745   &  0.0469557 \\
 Lesotho                          & -0.4049     & -0.536814  &  0.385506  \\
 Liberia                          & -0.811382   & -0.878786  &  0.231922  \\
 Madagascar                       & -0.735995   & -1.00382   &  0.425281  \\
 Malawi                           & -0.775441   & -0.953094  &  0.229014  \\
 Mali                             & -0.290661   & -0.500998  &  0.425105  \\
 Moldova                          &  0.996866   &  1.03015   &  0.146392  \\
 Mozambique                       & -0.560905   & -0.888259  &  0.616949  \\
 Myanmar                          &  0.147347   &  0.111866  &  0.292677  \\
 Namibia                          &  0.0822124  & -0.127053  &  0.9082    \\
 Nepal                            & -0.185337   & -0.185715  &  0.416537  \\
 Niger                            & -1.06198    & -1.34356   &  0.259844  \\
 Nigeria                          & -0.0903742  & -0.0900875 &  0.537997  \\
 Pakistan                         &  0.142732   &  0.146731  &  0.516279  \\
 Peru                             &  0.191104   &  0.310998  &  0.741152  \\
 Philippines                      &  0.694071   &  0.820905  &  0.342544  \\
 Rwanda                           & -0.812009   & -0.992644  &  0.248862  \\
 Senegal                          & -0.174405   & -0.318213  &  0.534392  \\
 Sierra Leone                     & -0.622954   & -0.862833  &  0.37829   \\
 Swaziland                        &  0.179458   &  0.0660039 &  0.433119  \\
 Tajikistan                       &  0.959175   &  1.04378   &  0.173927  \\
 Tanzania                         & -0.57875    & -0.789753  &  0.37443   \\
 Timor                            & -0.685584   & -0.904893  &  0.515151  \\
 Togo                             & -0.133553   & -0.308391  &  0.433274  \\
 Uganda                           & -0.847543   & -1.01354   &  0.241416  \\
 Zambia                           & -0.479619   & -0.764645  &  0.619943  \\
 Zimbabwe                         & -0.127609   & -0.444624  &  0.722577  \\
\bottomrule
\end{tabular}
\caption{This table summarizes the wealth statistics of the countries used during the experiments. Here, the average, median, and variance of wealth index is provided for each country.}
\label{tab:countries}
\end{table}

\subsection{Neural Network Development}
\begin{figure*}
    \begin{subfigure}{.99\textwidth}
        \centering
        \includegraphics[width=6in]{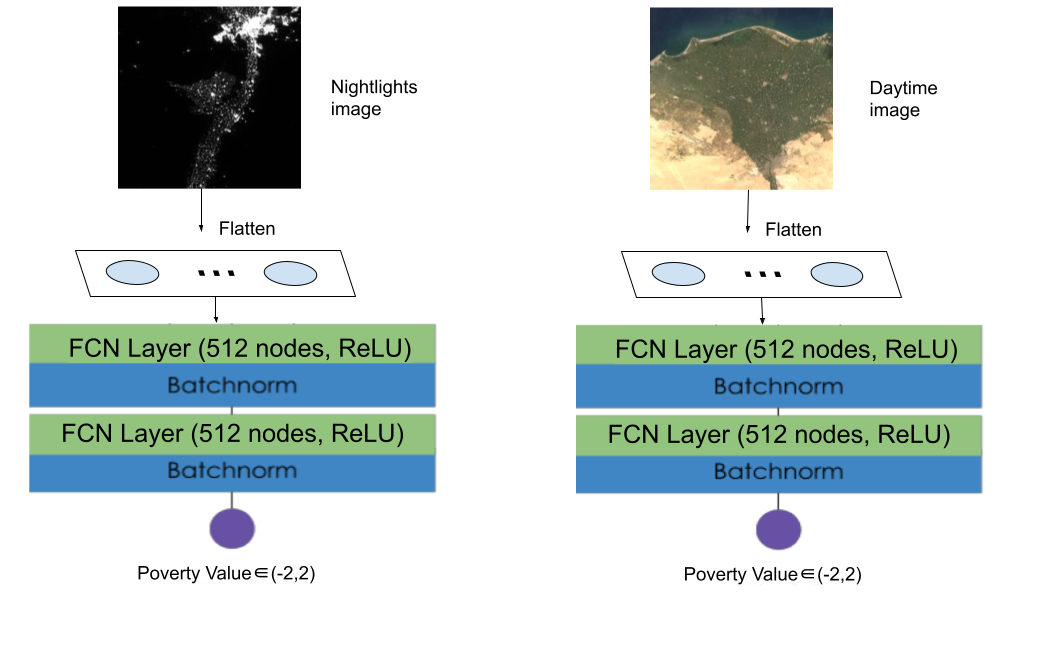}
    \end{subfigure}
    \caption{Architecture of the neural network used in experiments}
    \label{model-arch}
\end{figure*}

I trained multiple DNNs to predict the wealth index of a region (representing the poverty level).The DNNs had a single output each. Firstly, I flattened the input image, and then passed it through two multi-layer perceptrons, each with 512 nodes before finally outputting a wealth index score. The same architecture is used for daytime and nighttime images, as shown in Figure \ref{model-arch}.

An SGD optimizer with momentum \cite{sgd} is used with a learning rate of $10^{-7}$ for all experiments on nighttime imagery and a learning rate of $10^{-9}$ for all experiments on daytime imagery. A momentum value of $0.9$ was used for all experiments. I trained the DNNs using mean squared error (MSE) loss, as shown in Equation \ref{eqn:mse_loss}.

To ensure fair comparison across trials and experiments, I hold all hyperparameters and training procedures constant, with the exception of learning rate which differed for nighttime and daytime imagery as detailed above. However, within all networks trained on daytime images and all networks trained on nighttime images, the learning rate was held constant. Furthermore, all networks were trained for the 10 epochs. Because I empirically determined run-to-run variance for the same experiment to be low, I only ran each experiment once.

I developed the networks using Pytorch version 1.8 \cite{nineteen} using a single NVIDIA Tesla V100 graphics processing unit for training and evaluation.

\begin{equation}
    \label{eqn:mse_loss}
    L = \frac{1}{n} \sum_{i=1}^n (Y_i - \hat{Y}_i)^2
\end{equation}

\subsection{Checkpoint Selection}
All networks were trained on images only from the train split (See \emph{Dataset} section). After training, a checkpoint was selected based on what had the smallest RMSE value on the tune set, which consisted of images only from the tune split. Finally, using that checkpoint, inference was run on the test set, which consisted of images only from the test split. Only test set performance is reported.

\subsection{Evaluation and Statistical Analysis}
My main evaluation metric used in the analysis is root mean squared error (RMSE) which is one of the most commonly used metrics to assess performance on regression tasks \cite{rmse}. The goal is to minimize RMSE values. 

Confidence intervals (CI) for all evaluation metrics are calculated using the non-parametric bootstrap method with n=1000 permutations at the example level. 95\% confidence intervals are reported for all root mean squared errors (RMSE).

\section{Experiments}
To test my hypothesis that deep NNs can be trained to leverage satellite imagery to accurately identify the poverty level (wealth index) of the region along with several sub-hypotheses, I designed and executed numerous experiments. The experiments are aimed at evaluating the following:

\begin{enumerate}[label=(\alph*)]
    \item How well can a deep neural network predict the poverty level of a region given an overhead nighttime satellite image with a limited dataset?
    \item How well can a deep neural network predict the poverty level of a region given an overhead daytime satellite image with a limited dataset?
    \item What is the impact of data quantity (i.e. more or less data) on the representational power and overall accuracy of a deep neural network trained to predict the poverty level of a region given an overhead nighttime satellite image?
    \item What is the impact of data quantity (i.e. more or less data) on the representational power and overall accuracy of a deep neural network trained to predict the poverty level of a region given an overhead daytime satellite image?
    \item What is the impact of data augmentation on the representational power and overall accuracy of a deep neural network trained to predict the poverty level of a region given an overhead nighttime satellite image? How does data augmentation compare with a larger dataset?
    \item What is the impact of data augmentation on the representational power and overall accuracy of a deep neural network trained to predict the poverty level of a region given an overhead nighttime daytime image?
    \item How well can a deep neural network trained on a single continent’s data generalize to data from other continents?
\end{enumerate}

\begin{figure}[h!]
    \begin{subfigure}{.47\textwidth}
        \centering
        \includegraphics[width=3in]{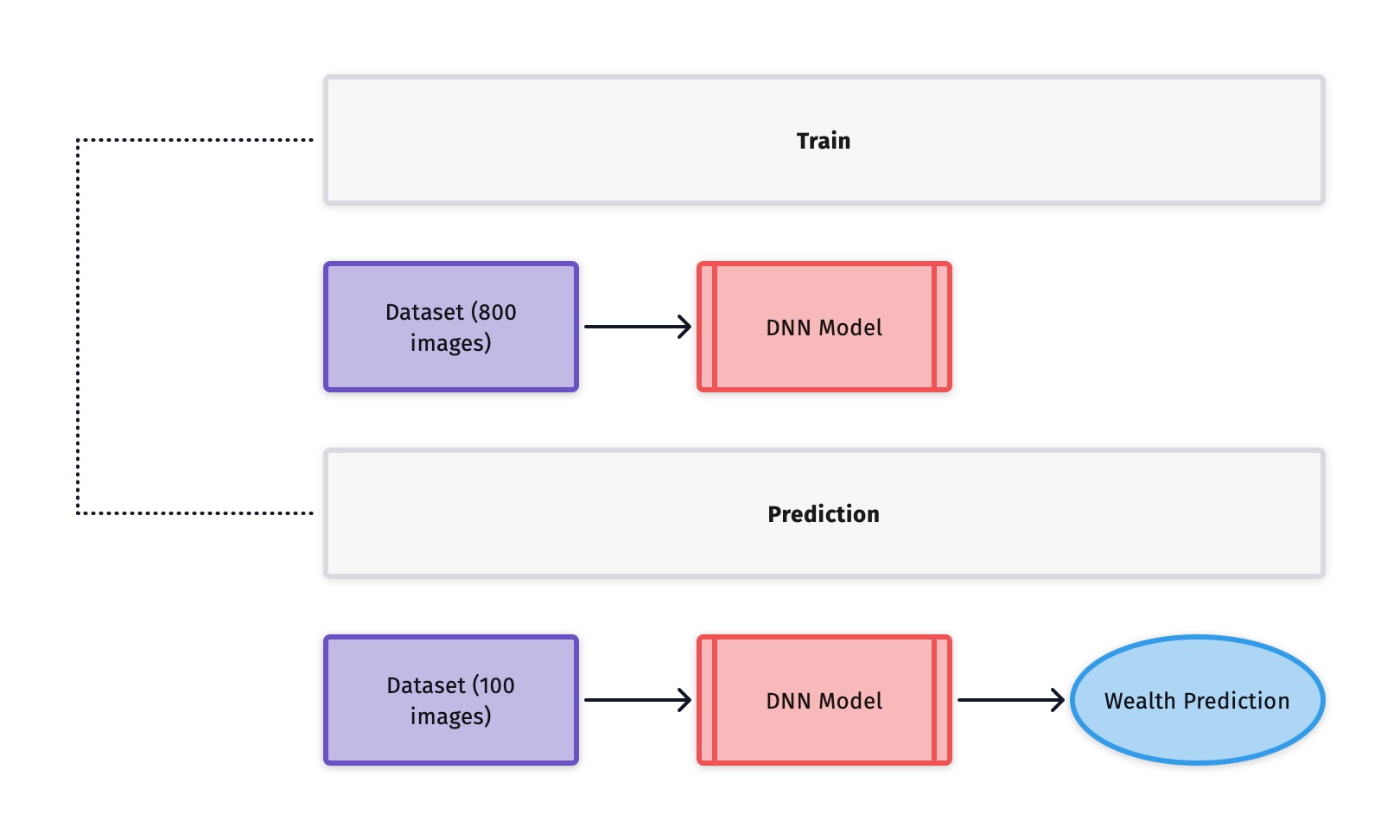}
        \caption{Experiment setup for baseline experiments.}
        \label{fig:exp_baseline}
    \end{subfigure}
        \begin{subfigure}{.47\textwidth}
        \centering
        \includegraphics[width=3in]{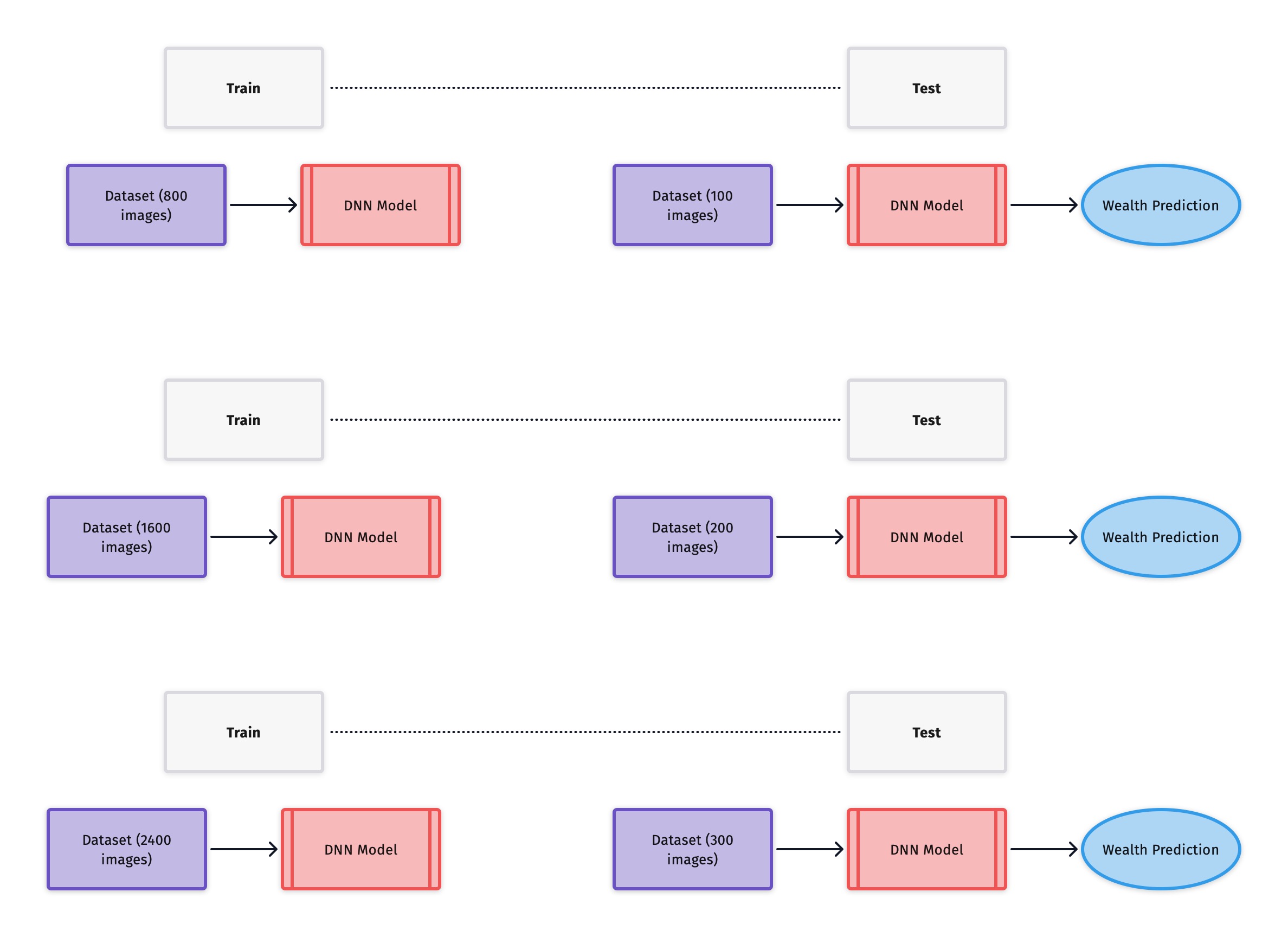}
        \caption{Experiment setup for data quantity experiments.}
        \label{fig:exp_quantity}
    \end{subfigure}
    \begin{subfigure}{.47\textwidth}
        \centering
        \includegraphics[width=3in]{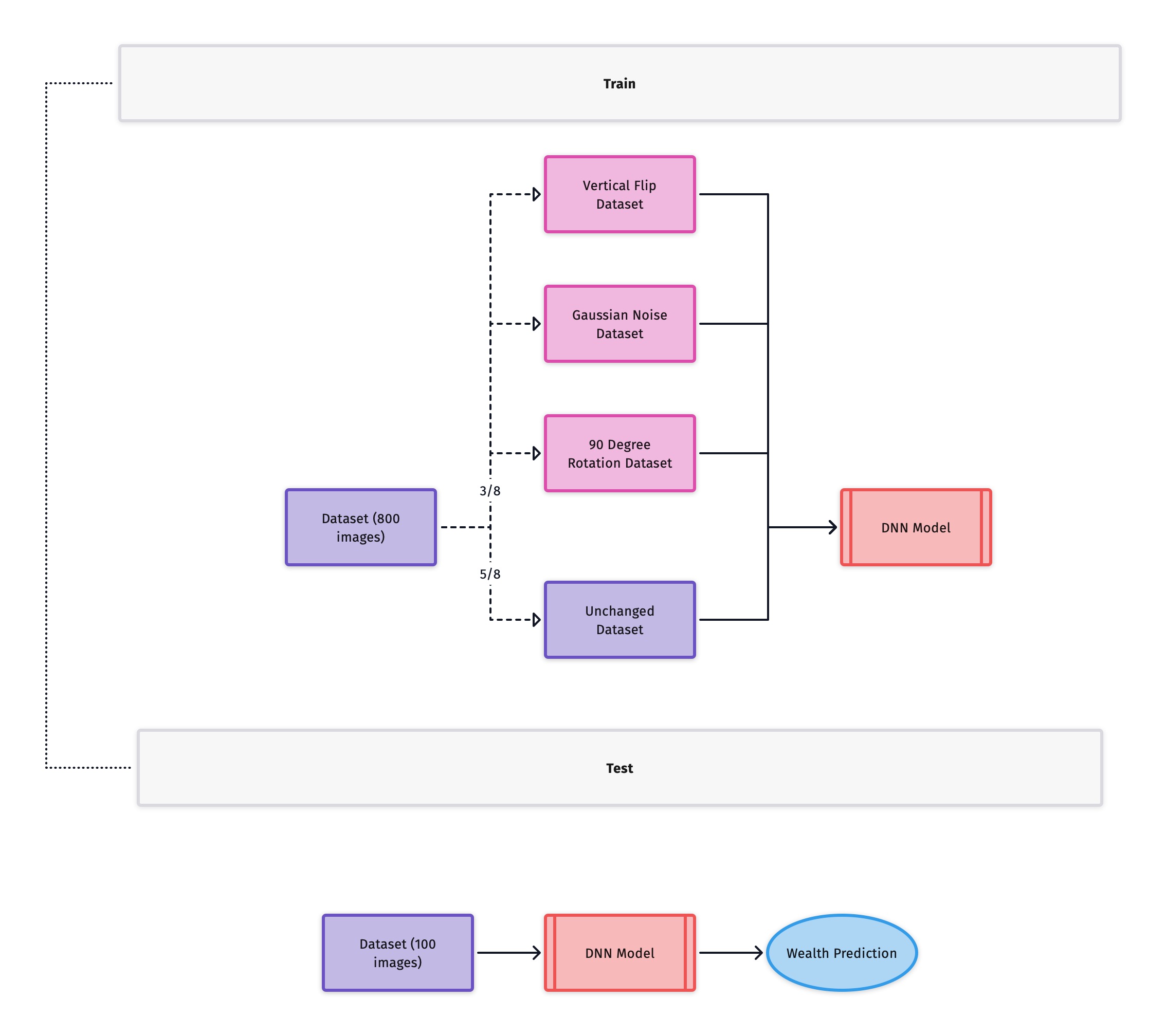}
        \caption{Experiment setup for data augmentation experiments.}
        \label{fig:exp_augmentation}
    \end{subfigure}
        \begin{subfigure}{.47\textwidth}
        \centering
        \includegraphics[width=3in]{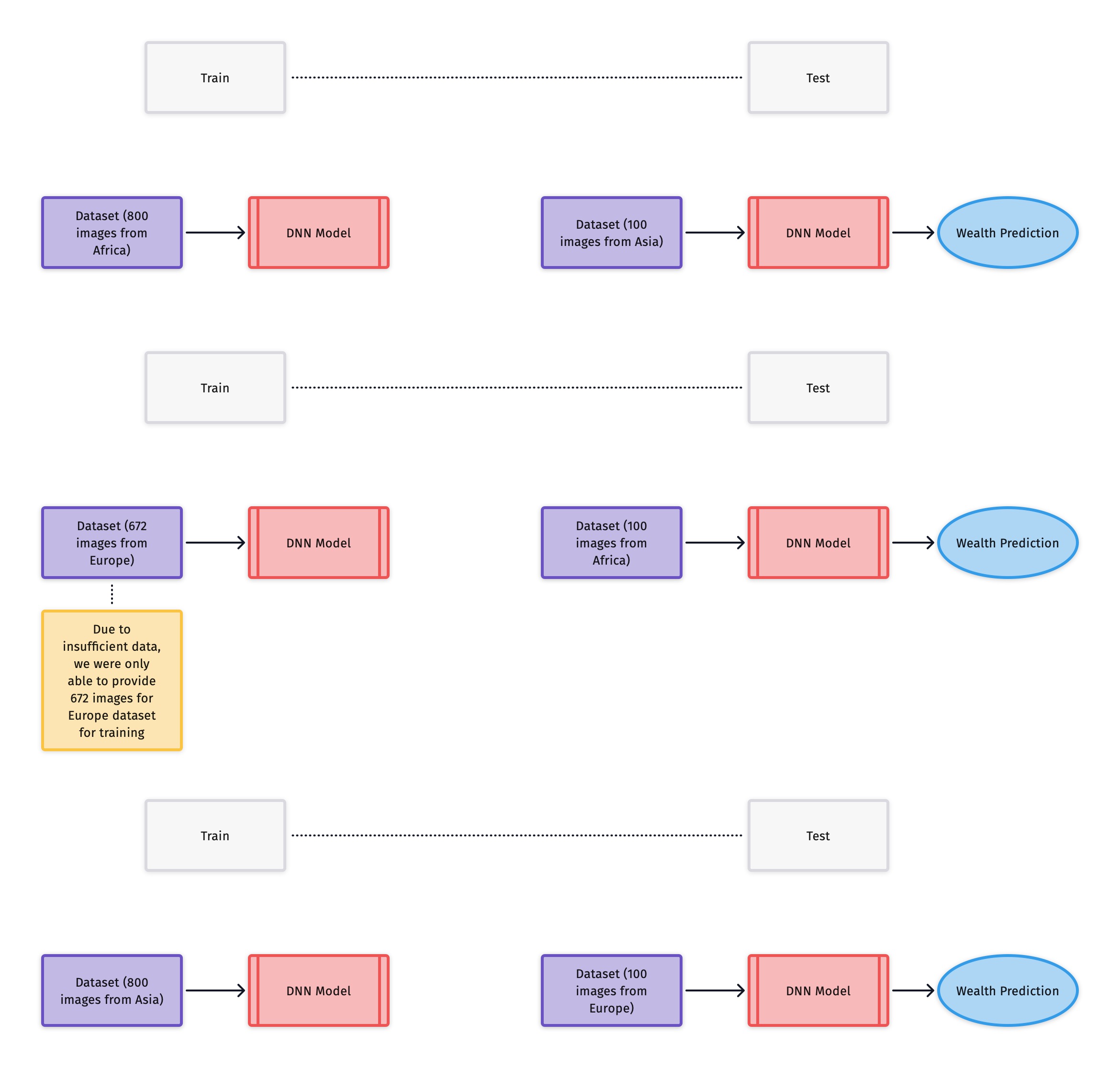}
        \caption{Experiment setup for generalization to unseen continents experiments.}
        \label{fig:exp_generalization}
    \end{subfigure}
    \caption{The four experiment setups presented in this work.}
    \label{fig:experiment_setup}
\end{figure}

See Figure \ref{fig:experiment_setup} for a schematic of all experiments run.

\subsection{Baseline: Using a Limited Dataset of Daytime and Nighttime Satellite Images}
To evaluate my top level hypothesis and establish a baseline for the remaining experiments, I ran experiments evaluating how well a deep neural network can predict the poverty level of a region based on a single daytime or nighttime satellite image, with a limited dataset. I restricted my training dataset to 800 images, and evaluated my network on 100 images, each randomly chosen from their respective splits (see \emph{Dataset} under \emph{Methods}). See Figure \ref{fig:exp_baseline} for an illustration of the experiment. I ran one experiment using daytime images, and one experiment using nighttime images and results are shown in Table \ref{tab:baseline}.

\subsection{Impact of Data Quantity}

Secondly, I wanted to better understand the importance of data quantity on the performance of the network for both daytime and nighttime image regression. It is a widely known and accepted fact that deep neural networks require a lot of data, and performance is often strongly correlated with data quantity \cite{data}. However, different tasks often require different amounts of data depending on the characteristics and difficulty of the tasks and the data available \cite{data2}. To evaluate the impact of data quantity on the representational power and overall accuracy of the DNN in the task of poverty prediction, I trained deep DNNs with the following amount of data:

\begin{enumerate}

    \item 800 images
    \item 1600 images
    \item 2400 images

\end{enumerate}

I ran the previous experiments once using daytime images and once using nighttime images for a total of 6 experiments. See Figure \ref{fig:exp_quantity} for an illustration of the experiment. Results are presented in Table \ref{tab:quantity}.

\subsection{Impact of Data Augmentation}

For deep learning tasks in the low data regime where obtaining labeled data is expensive or infeasible, a technique called data augmentation is often used to enlarge the dataset \cite{aug}. Existing works have explored data augmentation for satellite imagery \cite{sat-aug} with respectable results. To evaluate the impact of data augmentation on the representational power and overall accuracy of the DNNs for this specific task, I trained DNNs with various amounts of data augmentation, for both daytime images and nighttime images. 

The specific augmentations I used were random flipping (180 degree rotation followed by a mirror across the vertical axis through the center of the image), random rotations (90 degrees counter-clockwise), and random Gaussian noise with a mean of zero and standard deviation of 2.5 \cite{noise}. See Figure \ref{fig:exp_augmentation} for an illustration of the experiment. Results for experiments with data augmentation for both daytime and nighttime satellite imagery are presented in Table \ref{tab:augmentation}.

\subsection{Generalization to Unseen Continents}

While DNNs have been shown to match and even surpass human performance in various tasks ranging to medical image diagnosis \cite{cxr} to emotion recognition \cite{emotion}, a major issue these networks face is inability to generalize to data not seen during training. Thus, to robustly evaluate a performance of a DNN, it is important to assess its performance on unseen, "new" data to see how well it generalizes. For example, a neural network trained to distinguish between normal and abnormal chest X-rays, needs to be evaluated on its performance in the presence of new patient populations and new diseases \cite{Nabulsi}.

In the specific task of poverty prediction from satellite imagery, in order to robustly evaluate DNNs, it is similarly important to assess performance on "new" data. I explore this generalization by defining "new" data as data from a separate continent. To evaluate how well a DNN trained on satellite images from a single continent to predict poverty can generalize to another continent, I ran three experiments. For each experiment, I picked a continent at random to train on and a separate continent to evaluate on. Specifically, the three experiments are:

\begin{enumerate}

    \item Train on data from Africa, evaluate on data from Asia
    \item Train on data from Europe, evaluate on data from Africa
    \item Train on data from Asia, evaluate on data from Europe.

\end{enumerate}

Note that these experiments are only conducted on nighttime images because of the better performance. See Figure \ref{fig:exp_generalization} for an illustration of the experiment. Results of the above experiments can be found in Table \ref{tab:continent}.

\section{Results}
Results for the experiments detailed above are presented in Tables 2-5, and detailed in the respective sections below.

\subsection{Baselines}
For the task of predicting poverty, a DNN trained on nighttime images achieves a RMSE of 1.222 (95\% CI: 1.033-1.423) and a DNN trained on daytime images achieves a RMSE of 1.611 (95\% CI: 1.399-1.832). In both daytime and nighttime experiments, each DNN was trained on 800 images and evaluated on 100 images, each chosen from their respective data splits. Results are presented in Table \ref{tab:baseline}.

\begin{table}[]
\centering
\begin{tabular}{@{}ll@{}}
\toprule
Experiment Name & RMSE (95\% CI)                \\ \midrule
Baseline Night  & 1.222 (1.033-1.423)           \\
Baseline Day    & 1.611 (1.399-1.832)           \\ \bottomrule
\end{tabular}
\caption{This table summarizes the results of the model when simple day and night images were used to train and test the model. To evaluate the performance of the model 800 train images and 100 test images were used.}
\label{tab:baseline}
\end{table}

\subsection{Data Quantity and Data augmentation}
Results for experiments on data quantity and data augmentation are presented in Table \ref{tab:quantity} and Table \ref{tab:augmentation}, respectively. A DNN trained on nighttime images achieves RMSE's of 1.222 (95\% CI: 1.033-1.423), 1.132 (95\% CI: 0.963-1.354), and 1.106 (95\% CI: 0.956-1.283) when trained on 800, 1600, and 2400 images respectively.

A DNN trained on daytime images achieves RMSE's of 1.611 (95\% CI: 1.399-1.832), 1.459 (95\% CI: 1.275-1.633), and 1.477 (95\% CI: 1.203-1.894) when trained on 800, 1600, and 2400 images respectively. 

The same evaluation dataset of 100 images was used across all experiments for each of daytime and nighttime experiments for consistent comparisons.

\begin{table}[]
\centering
\begin{tabular}{@{}ll@{}}
\toprule
Experiment Name & RMSE (95\% CI)                 \\ \midrule
Night, 800 images  & 1.222 (1.033-1.423)         \\
Night, 1600 images  & 1.132 (0.963-1.354)        \\
Night, 2400 images  & 1.106 (0.956-1.283)        \\
Day, 800 images  & 1.611 (1.399–1.832)           \\
Day, 1600 images  & 1.459 (1.275-1.633)          \\
Day, 2400 images    & 1.477 (1.203-1.894)        \\ \bottomrule
\end{tabular}
\caption{This table summarizes the results of the model when the amount of data used to train the model was varied.}
\label{tab:quantity}
\end{table}

\begin{table}[]
\centering
\begin{tabular}{@{}ll@{}}
\toprule
Experiment Name & RMSE (95\% CI)                    \\ \midrule
Baseline Night  & 1.222 (1.033-1.423)               \\
Night with Augmentation  & 0.956 (0.872-1.040)      \\
Baseline Day  & 1.611 (1.399-1.832)                 \\
Day with Augmentation  & 1.463 (1.221-1.705)        \\ \bottomrule
\end{tabular}
\caption{This table summarizes the results of the model when the data used to train the model was partially augmented.}
\label{tab:augmentation}
\end{table}

\subsection{Performance in the Setting of Data from Unseen Continents}
Table \ref{tab:continent} shows results for my generalization experiments. A DNN trained on data from Africa and evaluated on data from Asia achieves RMSE of 0.926 (95\% CI 0.814-1.031). Additionally, a DNN trained on data from Europe and evaluated on data from Africa achieves RMSE of 3.182 (95\% CI 1.981-4.321). Lastly, a DNN trained on data from Asia and evaluated on data from Europe achieves RMSE of 1.156 (95\% CI 1.014-1.285).

All three experiments here only used nighttime images. For all three experiments, the same evaluation set of 100 images was used.

\begin{table}[]
\centering
\begin{tabular}{@{}llll@{}}
\toprule
Experiment Name & Train Continent & Test Continent & RMSE (95\% CI)      \\ \midrule
Experiment 1    & Africa          & Asia           & 0.926 (0.814-1.031) \\
Experiment 2    & Europe*         & Africa         & 3.182 (1.981-4.321) \\
Experiment 3    & Asia            & Europe         & 1.156 (1.014-1.285) \\ \bottomrule
\end{tabular}

\footnotesize{*The train set for Europe was slightly smaller due to insufficient data with 672 images compared to 800 images for the rest of the experiments.}
\caption{This table summarizes the result of the model when separate continents were used to train and test the model. For this experiment, only night time images were used.}
\label{tab:continent}
\end{table}

\subsection{Distributional Shifts between Countries/Continents}
As expected, there was a fairly significant data distribution shift across countries and continents. Table \ref{tab:countries} presents key statistics regarding wealth indices on the country level, where the distribution shift is dramatic.

Median wealth indices for each country range from -1.238 in Ethiopia to 1.279 in Armenia. Variance of the wealth index also differed dramatically across countries, ranging from 0.908 in Namibia to 0.0380 in Armenia. Variance in the wealth index is a proxy for wealth inequality within a country where higher variance indicates greater inequality.

\section{Discussion}
In this paper, I developed a multitude of deep neural networks under various circumstances with the goal of predicting a region's poverty level from overhead satellite images. Both daytime and nighttime satellite images were used as input. Furthermore, different deep learning methods were utilized in order to better understand the task. I designed and ran experiments to determine the impact of data quantity and data augmentation on the accuracy of the DNNs. Lastly, I designed and ran experiments to evaluate how well DNNs generalize to data from continents that were absent in the training set.

For the task of poverty prediction, the results show that a DNN can be trained to estimate a region's poverty from a satellite image fairly accurately, with RMSE values of only slightly above 1. Furthermore, from Table \ref{tab:baseline}, I see that a DNN can better estimate a region's povery from the nighttime satellite image rather than a daytime satellite image. This is expected, given that a nighttime satellite image shows the amount of light in a region, which is often correlated with an active economy (see Figure \ref{fig:night_day} for sample daytime and nighttime satellite images). The better performance on nighttime imagery also shows that a nighttime satellite image holds more information regarding a region's poverty level than a daytime satellite image. However, that is not to say that a daytime satellite image contains no clues regarding the  wealth of a given region as the performance of a DNN trained on daytime images has RMSE as low as 1.477, which is surprisingly accurate. The main takeaway here is that both daytime and nighttime satellite imagery can be used to effectively estimate the poverty level of a region.

Experiments on data quantity and data augmentation showed that, as expected, DNN performance improved with more data for both daytime and nighttime satellite imagery (either through data augmentation methods or through obtaining more organic data). Significant increases in performance are seen when the data is augmented from 800 to 1600, decreasing by 0.09 RMSE (7.36\%) and 0.152 RMSE (9.44\%) for nighttime and daytime networks respectively. Performance started to stagnate past a training dataset size of 1600. A decrease of 0.026 RMSE (2.30\%) was observed for the nighttime network and an increase of 0.018 (1.23\%) was observed for the daytime network, indicating that more data is not necessarily useful past 1600 images. While seeing the trend of stagnating performance with continuous training data increases is common across deep learning, the number of images this happens is normally much higher than 1600 images. This indicates that training DNNs for poverty prediction from satellite imagery may not need as much data. Furthermore, this also indicates that the at 1600 images, the constraint is no longer amount of data, but rather the representational power of the DNN. For more improvements, a more complicated architecture (i.e. more layers) is likely needed.

Similar to training data increases, train-time data augmentation provided a comparable a benefit. For the nighttime network, RMSE decreased by 0.266 (21.8\%). For the daytime network, RMSE decreased by 0.148 (9.19\%). These results demonstrate that in addition to obtaining more labeled data, using data augmentation methods are also beneficial. Similarly to the results obtained by increasing training data size, data augmentation seems to provide more of a boost to nighttime networks. Not only does data augmentation synthetically increase the dataset size, but it also serves as an important regularizer.

Interestingly, data augmentation seemed to provide a larger benefit to the nighttime network than increasing the development dataset size (achieving a decrease of 0.266 RMSE compared to 0.116 RMSE). For the daytime networks, the performance change is comparable (roughly 0.15 decrease in RMSE). The fact that performance increases are more dramatic in the nighttime images than the daytime images indicates that the DNN's representational power can be increased by having a more complex architecture, specifically for daytime images. For nighttime images, performance may not have fully saturated with the current network. 

The results of the generalization experiments are also insightful. Two of the three experiments show that the DNN can generalize well to data from unseen continents. A DNN trained on data from only Africa achieves an RMSE of 0.926 on data from Asia, better than what the general-purpose model performs on a general-purpose evaluation set. This demonstrates great generalization power of the DNN. A similar trend is seen in the DNN trained on data from Asia and evaluated on data from Europe, achieving an RMSE 1.156. However, a completely different result is observed on the network trained on Europe and evaluated on Africa, where an RMSE of 3.182 is noted. While the train set in this experiment is slightly smaller (672 images compared to 800 for the prior two), this smaller dataset likely is not sufficient to explaining the huge performance drop, indicating that data in Europe is very different than data in Africa.

As alluded to earlier, in this work, a simple neural network architecture was used in this paper, consisting of only two layers of 512 nodes each (see Figure \ref{model-arch}). This was done by design, as if a more complex architecture was used,  it would be harder (or even infeasible) to saturate the network's performance based on its representational power with the the available data. Furthermore, the goal of this work was not to develop the most accurate network to estimate poverty level, but rather evaluate how  different deep learning techniques impact performance. Because the DNN architecture and hyperparameters were held constant across experiments, the impact of each technique could be better assessed.

\section{Limitations and Future Work}
While this work presents an important advancement in the intersection of computer vision and poverty prediction from satellite imagery, there are a few limitations worth discussing. First of all, the data is imperfect by nature. The way I paired wealth indices with satellite imagery is not exact, as the wealth index is given at a specific coordinate, but satellite images are somewhat sparse. The closest daytime and nighttime satellite image to the coordinates of the wealth index was chosen, but sometimes this point was not as close as ideal. However, since neighboring cities tend to have similar wealth indices, this may not be as big of an issue in practice. Furthermore, only a subset of the available data was used. while this was necessary to better analyze the differences and impact of each deep learning technique, using more data would likely result in more result networks.

On the modeling front, only a single architecture was used for all experiments, and no extensive hyperparameter search was conducted due to computational constraints. For example, a convolutional neural network (CNN) would be expected to perform significantly better. In practice, this likely would have resulted in improved performances, but this was not the main goal of this work. An additional limitation is that while both increasing data quantity and data augmentation were both assessed, the conjunction was never tried, and could lead to insightful results.

\begin{figure}[t]
    \centering
    \includegraphics[width=6.2in]{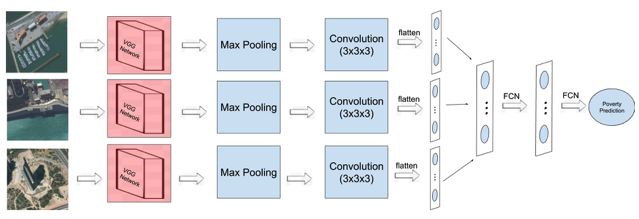}
    \caption{Architecture of a multi-image model CNN.}
    \label{fig:Multi_Image_CNN}
\end{figure}

While the results presented in this work are exciting, there is still plenty of work to be done in order to realize the impact on a global scale. Two main ideas immediately stem from this work. The first is that a network can be trained to take in as input both a daytime and a nighttime satellite image to reap the benefits of both. Furthermore, since oftentimes multiple satellite images of a single city are often available, these can all be provided to a network as input (see Figure \ref{fig:Multi_Image_CNN}.1

\section{Conclusion}
The goal of this research was to examine a novel approach to poverty prediction using both daytime and nighttime satellite images. While there is more to do, I think my work can provide a basis for poverty prediction as well as show its potential. Identifying regions of poverty is the first step on its reduction and I believe my work has made a contribution.

\section{Acknowledgements}
I would like to acknowledge the contributions of Zaid Nabulsi for his guidance and mentorship in tackling this problem, as well as his help in exploring previous research efforts. I’d also like to acknowledge the contributions of Professor Stefano Ermon from Stanford and his Sustainability and Artificial Intelligence Lab for inspiring this research and early efforts.

{\small
\bibliographystyle{ieee}
\bibliography{egbib}

\begin{thebibliography}{10}\itemsep=-1pt

\bibitem{sixteen}
\url{https://landsat.gsfc.nasa.gov/.}
\newblock Accessed: 2021-09-12.

\bibitem{seventeen}
\url{https://earthdata.nasa.gov/earth-observation-data/near-real-time/download-nrt-data/viirs-nrt.}
\newblock Accessed: 2021-09-12.

\bibitem{eighteen}
\url{https://data.worldbank.org/.}
\newblock Accessed: 2021-09-12.

\bibitem{fifteen}
Google earth engine.
\newblock \url{https://developers.google.com/earth-engine}.
\newblock Accessed: 2021-09-12.

\bibitem{albert2017using}
A.~Albert, J.~Kaur, and M.~C. Gonzalez.
\newblock Using convolutional networks and satellite imagery to identify
  patterns in urban environments at a large scale.
\newblock In {\em Proceedings of the 23rd ACM SIGKDD International Conference
  on Knowledge Discovery and Data Mining}, pages 1357--1366. ACM, 2017.

\bibitem{audebert2017joint}
N.~Audebert, B.~Le~Saux, and S.~Lef{\`e}vre.
\newblock Joint learning from earth observation and openstreetmap data to get
  faster better semantic maps.
\newblock In {\em EARTHVISION 2017 IEEE/ISPRS CVPR Workshop. Large Scale
  Computer Vision for Remote Sensing Imagery}, 2017.

\bibitem{rmse}
T.~Chai and R.~R. Draxler.
\newblock Root mean square error (rmse) or mean absolute error (mae)? –
  arguments against avoiding rmse in the literature.
\newblock {\em Geoscientific Model Development}, 7(3):1247--1250, 2014.

\bibitem{sat-aug}
P.~R. Dave and H.~A. Pandya.
\newblock Satellite image classification with data augmentation and
  convolutional neural network.
\newblock In {\em Lecture Notes in Electrical Engineering}, Lecture notes in
  electrical engineering, pages 83--92. Springer Singapore, Singapore, 2020.

\bibitem{one}
{dpicampaigns} and {fangweizhao}.
\newblock Goal 1: End poverty in all its forms everywhere.
\newblock \url{https://www.un.org/sustainabledevelopment/poverty/.}
\newblock Accessed: 2021-09-27.

\bibitem{five}
J.~Han, D.~Zhang, G.~Cheng, N.~Liu, and D.~Xu.
\newblock Advanced deep-learning techniques for salient and category-specific
  object detection: A survey.
\newblock {\em IEEE Signal Processing Magazine}, 35(1):84--100, 2018.

\bibitem{Jean790}
N.~Jean, M.~Burke, M.~Xie, W.~M. Davis, D.~B. Lobell, and S.~Ermon.
\newblock Combining satellite imagery and machine learning to predict poverty.
\newblock {\em Science}, 353(6301):790--794, 2016.

\bibitem{eight}
M.~I. Jordan and T.~M. Mitchell.
\newblock Machine learning: Trends, perspectives, and prospects.
\newblock {\em Science (New York}, 349(6245):255, 2015.

\bibitem{kim2006satellite}
S.-W. Kim, A.~Heckel, S.~McKeen, G.~Frost, E.-Y. Hsie, M.~Trainer, A.~Richter,
  J.~Burrows, S.~Peckham, and G.~Grell.
\newblock Satellite-observed us power plant nox emission reductions and their
  impact on air quality.
\newblock {\em Geophysical Research Letters}, 33(22), 2006.

\bibitem{emotion}
S.~Li and W.~Deng.
\newblock Deep facial expression recognition: A survey.
\newblock {\em IEEE Transactions on Affective Computing}, page 1–1, 2020.

\bibitem{four}
T.-Y. Lin, M.~Maire, S.~Belongie, L.~Bourdev, R.~Girshick, J.~Hays, P.~Perona,
  D.~Ramanan, C.~L. Zitnick, and P.~Dollár.
\newblock Microsoft coco: Common objects in context, 2015.

\bibitem{fourteen}
M.~Mirza and S.~Osindero.
\newblock Conditional generative adversarial nets.
\newblock Nov. 2014.

\bibitem{Nabulsi}
Z.~Nabulsi, A.~Sellergren, S.~Jamshy, C.~Lau, E.~Santos, A.~P. Kiraly, W.~Ye,
  J.~Yang, R.~Pilgrim, S.~Kazemzadeh, J.~Yu, S.~R. Kalidindi, M.~Etemadi,
  F.~Garcia-Vicente, D.~Melnick, G.~S. Corrado, L.~Peng, K.~Eswaran, D.~Tse,
  N.~Beladia, Y.~Liu, P.-H.~C. Chen, and S.~Shetty.
\newblock Deep learning for distinguishing normal versus abnormal chest
  radiographs and generalization to two unseen diseases tuberculosis and
  {COVID-19}.
\newblock {\em Sci. Rep.}, 11(1):15523, Sept. 2021.

\bibitem{nineteen}
A.~Paszke, S.~Gross, F.~Massa, A.~Lerer, J.~Bradbury, G.~Chanan, T.~Killeen,
  Z.~Lin, N.~Gimelshein, L.~Antiga, A.~Desmaison, A.~K{\"o}pf, E.~Yang,
  Z.~DeVito, M.~Raison, A.~Tejani, S.~Chilamkurthy, B.~Steiner, L.~Fang,
  J.~Bai, and S.~Chintala.
\newblock {PyTorch}: An imperative style, high-performance deep learning
  library.
\newblock Dec. 2019.

\bibitem{nips_2017_workshop_stefano}
A.~Perez, C.~Yeh, G.~Azzari, M.~Burke, D.~Lobell, and S.~Ermon.
\newblock Poverty prediction with public landsat 7 satellite imagery and
  machine learning.
\newblock 11 2017.

\bibitem{eleven}
A.~Perez, C.~Yeh, G.~Azzari, M.~Burke, D.~Lobell, and S.~Ermon.
\newblock Poverty prediction with public landsat 7 satellite imagery and
  machine learning.
\newblock Nov. 2017.

\bibitem{cxr}
P.~Rajpurkar, J.~Irvin, K.~Zhu, B.~Yang, H.~Mehta, T.~Duan, D.~Ding, A.~Bagul,
  C.~Langlotz, K.~Shpanskaya, M.~P. Lungren, and A.~Y. Ng.
\newblock Chexnet: Radiologist-level pneumonia detection on chest x-rays with
  deep learning, 2017.

\bibitem{sgd}
S.~Ruder.
\newblock An overview of gradient descent optimization algorithms.
\newblock {\em arXiv preprint arXiv:1609.04747}, 2016.

\bibitem{noise}
E.~Rusak, L.~Schott, R.~S. Zimmermann, J.~Bitterwolf, O.~Bringmann, M.~Bethge,
  and W.~Brendel.
\newblock A simple way to make neural networks robust against diverse image
  corruptions, 2020.

\bibitem{six}
O.~Russakovsky, J.~Deng, H.~Su, J.~Krause, S.~Satheesh, S.~Ma, Z.~Huang,
  A.~Karpathy, A.~Khosla, M.~Bernstein, A.~C. Berg, and L.~Fei-Fei.
\newblock Imagenet large scale visual recognition challenge, 2015.

\bibitem{data}
M.-A. Schulz, B.~T.~T. Yeo, J.~T. Vogelstein, J.~Mourao-Miranada, J.~N. Kather,
  K.~Kording, B.~Richards, and D.~Bzdok.
\newblock Different scaling of linear models and deep learning in {UKBiobank}
  brain images versus machine-learning datasets.
\newblock {\em Nat. Commun.}, 11(1):4238, Aug. 2020.

\bibitem{aug}
C.~Shorten and T.~M. Khoshgoftaar.
\newblock A survey on image data augmentation for deep learning.
\newblock {\em J. Big Data}, 6(1), Dec. 2019.

\bibitem{nine}
N.~Srivastava, G.~Hinton, A.~Krizhevsky, I.~Sutskever, and R.~Salakhutdinov.
\newblock Dropout: A simple way to prevent neural networks from overfitting.
\newblock {\em Journal of Machine Learning Research}, 15(56):1929--1958, 2014.

\bibitem{data2}
V.~Sze, Y.-H. Chen, T.-J. Yang, and J.~Emer.
\newblock Efficient processing of deep neural networks: A tutorial and survey,
  2017.

\bibitem{two}
L.~S. Tusting, D.~Bisanzio, G.~Alabaster, E.~Cameron, R.~Cibulskis, M.~Davies,
  S.~Flaxman, H.~S. Gibson, J.~Knudsen, C.~Mbogo, F.~O. Okumu, L.~von Seidlein,
  D.~J. Weiss, S.~W. Lindsay, P.~W. Gething, and S.~Bhatt.
\newblock Mapping changes in housing in sub-saharan africa from 2000 to 2015.
\newblock {\em Nature}, 568(7752):391--394, Apr. 2019.

\bibitem{thirteen}
B.~Uzkent, E.~Sheehan, C.~Meng, Z.~Tang, M.~Burke, D.~Lobell, and S.~Ermon.
\newblock Learning to interpret satellite images in global scale using
  wikipedia.
\newblock May 2019.

\bibitem{wang2015deep}
J.~Wang, Q.~Qin, Z.~Li, X.~Ye, J.~Wang, X.~Yang, and X.~Qin.
\newblock Deep hierarchical representation and segmentation of high resolution
  remote sensing images.
\newblock In {\em Geoscience and Remote Sensing Symposium (IGARSS), 2015 IEEE
  International}, pages 4320--4323. IEEE, 2015.

\bibitem{wang2012poverty}
W.~Wang, H.~Cheng, and L.~Zhang.
\newblock Poverty assessment using dmsp/ols night-time light satellite imagery
  at a provincial scale in china.
\newblock {\em Advances in Space Research}, 49(8):1253--1264, 2012.

\bibitem{xia2018dota}
G.-S. Xia, X.~Bai, J.~Ding, Z.~Zhu, S.~Belongie, J.~Luo, M.~Datcu, M.~Pelillo,
  and L.~Zhang.
\newblock Dota: A large-scale dataset for object detection in aerial images.
\newblock In {\em Proc. CVPR}, 2018.

\bibitem{twelve}
M.~Xie, N.~Jean, M.~Burke, D.~Lobell, and S.~Ermon.
\newblock Transfer learning from deep features for remote sensing and poverty
  mapping.
\newblock Sept. 2015.

\bibitem{three}
C.~Yeh, A.~Perez, A.~Driscoll, G.~Azzari, Z.~Tang, D.~Lobell, S.~Ermon, and
  M.~Burke.
\newblock Using publicly available satellite imagery and deep learning to
  understand economic well-being in africa.
\newblock {\em Nat. Commun.}, 11(1):2583, May 2020.

\bibitem{seven}
S.~S.~A. Zaidi, M.~S. Ansari, A.~Aslam, N.~Kanwal, M.~Asghar, and B.~Lee.
\newblock A survey of modern deep learning based object detection models, 2021.

\end{thebibliography}
}

\end{document}